\documentclass[conference]{IEEEtran}
\usepackage{times}

\usepackage{multicol}
\usepackage[bookmarks=true]{hyperref}

\usepackage{hyperref}
\usepackage{amssymb}  
\usepackage{algorithm} 
\usepackage{graphicx}
\graphicspath{ {figs/} }
\usepackage{float}
\usepackage{
    amsmath,
    float,
    xcolor,
    marginnote,
    subcaption,
    pgfgantt,
    times,
    outlines,
    wrapfig,
    romannum,
    csquotes,
    bm,
    algorithm,
    algpseudocode,
}

\usepackage{lipsum}
\usepackage{soul,xcolor}

\setstcolor{red}
\usepackage{tabularx}
\usepackage{adjustbox}
\usepackage{graphicx}
\usepackage[doi=false,isbn=false,defernumbers=true,style=numeric,,backend=biber,maxbibnames=1000,sorting=none,firstinits=true]{biblatex}

\newcommand{\osnote}[1]{{}}
\newcommand{\rgnote}[1]{{}}
\newcommand{\jlnote}[1]{{}}
\newcommand{\jsnote}[1]{{}}
\newcommand{\mvnote}[1]{{}}
\newcommand{\todo}[1]{{}}
\newcommand{\assignTo}[1]{{}}
\newcommand{\revisit}[1]{{}}

\newcommand{\bs}{{\mathbf s}}
\newcommand{\ba}{{\mathbf a}}
\newcommand{\br}{{\mathbf r}}

\newcommand{\eg}{\textit{e}.\textit{g}. }
\makeatletter
\let\blx@rerun@biber\relax
\makeatother
\begin{document}

\title{Robust Multi-Modal Policies for Industrial Assembly via Reinforcement Learning and Demonstrations: A  Large-Scale Study}



\author{Jianlan Luo*\ \ Oleg Sushkov*\  \  Rugile Pevceviciute*\ \  Wenzhao  Lian\  \ Chang Su\  \\ \\   Mel Vecerik\ \ Ning Ye\  \ Stefan Schaal\  \ Jon Scholz \\ \\
Google X \\
DeepMind\\
*Equal contribuition}


%

\author{
\IEEEauthorblockN{Jianlan Luo*}
\IEEEauthorblockA{Google X\\
}\\   
\IEEEauthorblockN{Wenzhao Lian}
\IEEEauthorblockA{Google X \\
} \\
\IEEEauthorblockN{Ning Ye}
\IEEEauthorblockA{Google X\\
}
\and
\IEEEauthorblockN{Oleg Sushkov*}
\IEEEauthorblockA{DeepMind\\
}\\  
\IEEEauthorblockN{Chang Su}
\IEEEauthorblockA{Google X\\} \\
\IEEEauthorblockN{Stefan Schaal}
\IEEEauthorblockA{Google X\\}
\and
\IEEEauthorblockN{Rugile Pevceviciute*}
\IEEEauthorblockA{DeepMind\\}\\

\IEEEauthorblockN{ Mel Vecerik}
\IEEEauthorblockA{DeepMind} \\
\IEEEauthorblockN{Jon Scholz}
\IEEEauthorblockA{DeepMind\\}
}

\maketitle
\def\thefootnote{*}\footnotetext{These authors contributed equally }\def\thefootnote{}
\footnote{Email: \{jianlanluo, sushkov, rugile, wenzhaol, suchang, vec, nye, sschaal, jscholz\}@google.com}

\begin{abstract}
Over the past several years there has been a considerable research investment into learning-based approaches for tasks inspired by industrial manufacturing, but despite significant progress, these techniques have yet to be adopted by in the real-world.
We argue that it is the prohibitively large design space for Deep Reinforcement Learning (DRL), rather than algorithmic limitations \textit{per se}, that are truly responsible for this lack of adoption.  
Pushing these techniques into the industrial mainstream requires a paradigm which differs significantly from the academic mindset.
In this paper we define criteria for industry-oriented DRL, and perform a thorough comparison according to these criteria of one family of learning approaches, DRL from demonstration, against results of a professional industrial integrator on the recently established NIST assembly benchmark.
We explain the design choices, representing several years of investigation, which enabled our DRL system to consistently outperform the integrator's baseline in terms of both speed and reliability.
Finally, we conclude with a competition between our DRL system and a human on a challenge task of insertion into a randomly moving target.  
This study suggests that DRL is capable of outperforming not only established engineered approaches, but the human motor system as well, and that there remains significant room for improvement. Videos can be found on our project website:\url{https://sites.google.com/view/shield-nist}.




\end{abstract}


\section{INTRODUCTION}

Since the rise of deep-learning there has been an increasing interest in using learning-based methods to solve industrial manipulation tasks.
The two dominant families of industry-inspired tasks using deep-learning today are mixed bin-picking, which poses a significant perception challenge, and insertion, which emphasizes contact-rich control and multi-modal feedback.
While mixed-picking has seen some adoption in industry, the stigma of complexity, brittleness, and long training times for reinforcement-learning (RL) systems has hampered their adoption for industrial insertion.


\begin{figure}[t]
    \centering
   \includegraphics[width=9cm,keepaspectratio]{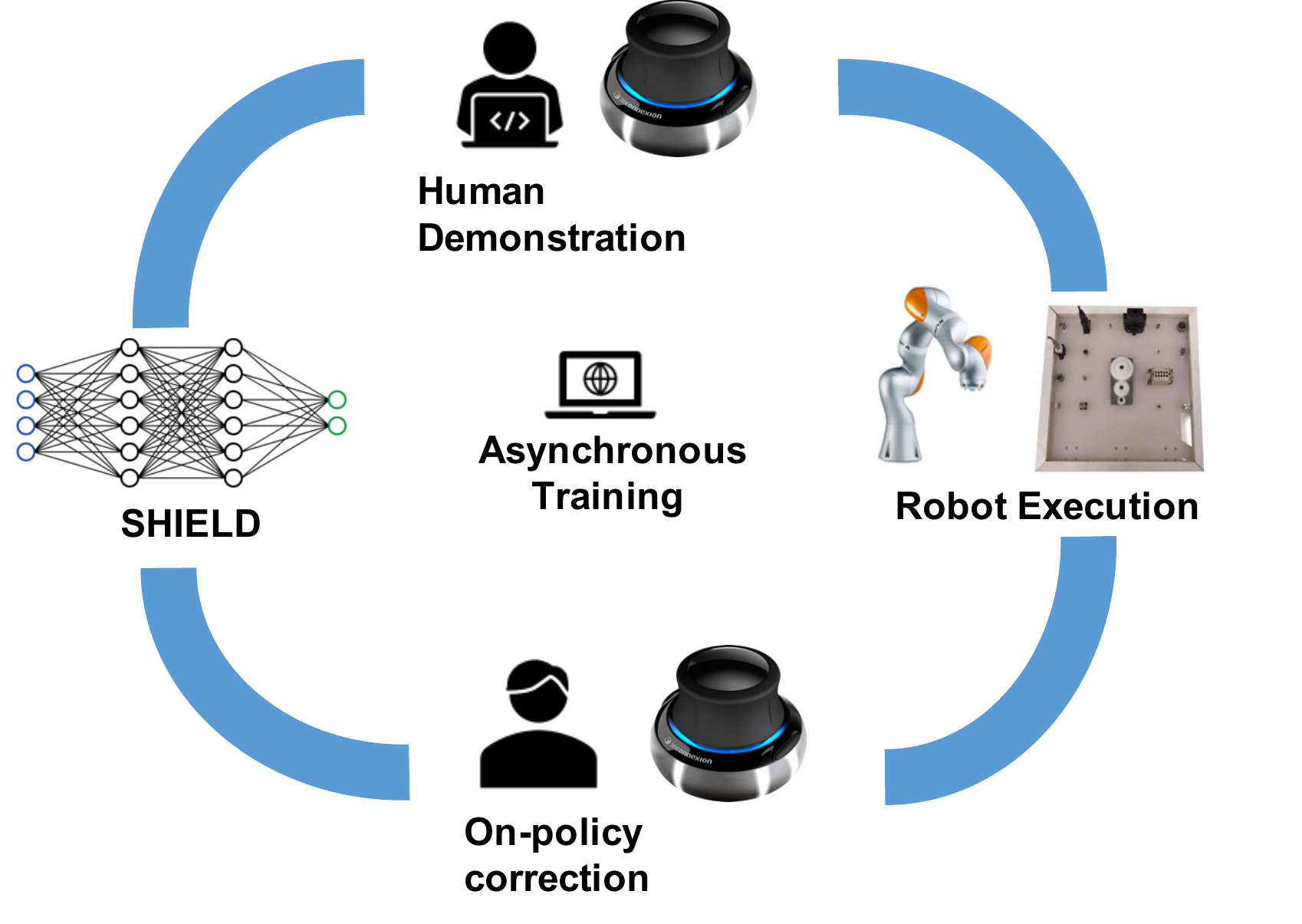}
    \caption{An overview of our method: (1) gather human demonstration data by teleportation, (2) robot executes its current policy from the latest snapshot, (3) human perform occasional on-policy correction if necessary, (4) asynchronous training of an improved DDPGfD agent:SHIELD.}
    \label{cover_photo}
\end{figure}
Instead, most industrial insertion robots today operate in constrained environments and execute sequences of manually pre-programmed force-control or vision-control primitives.
This approach imposes a barrier to providing higher levels of autonomy at a reasonable cost, since each manipulation skill requires significant time and effort to engineer.
As a result, entire sectors of low-volume manufacturing are precluded from accessing the benefits of robotic automation.

Deep reinforcement learning (DRL) offers the potential to autonomously train motor skills with little or no additional engineering effort. 
However, these methods have not been pushed to industry-acceptable standards, nor systematically evaluated against industry acceptable benchmarks.
Previous attempts to demonstrate learned insertion tasks \cite{Rajeswaran-RSS-18,vecerik2017leveraging,vecerik2019practical,luo2018deep,luo2019reinforcement,schoettler2020meta,inoue2017deep} have varied wildly in the task definition, action space, connector type, and sensor observations.
Fortunately, the recently introduced NIST assembly benchmark \cite{kimble2020benchmarking} provides an externally defined set of representative industrial assembly tasks with clearly defined metrics.

Inspired by the NIST benchmark, we argue that there are three key attributes for DRL to succeed in industrial robotic settings:\footnote{Safety is often cited as a fourth factor but is omitted from this list because, in the ``last-inch'' regimes we consider here, safety can be attained by an appropriate impedance/admittance controller.  The lack of compliant industrial robots is thus another significant obstacle to deploying DRL in the real world, but is not addressed by this work.}
\begin{enumerate}
    \item Efficient.  \jsnote{this is too solution-oriented, rephrase as a statement of the problem, ie it's gotta be fast to train.} It must be off-policy, and have a proper mechanism for injecting suitable priors into the learning process, such that one general RL algorithm can quickly adapt to new tasks by adapting task-specific priors.  In this paper we argue that human demonstration is a simple and natural way to provide such prior information. 
    \item Economical.  The algorithm must bring value to the automation process. This can be done by enabling solutions to problems that are infeasible with existing techniques while being simple enough to deploy in practice. 
    \item Evaluated. The resulting algorithm has to be evaluated extensively on industrial benchmarks using industry-relevant metrics such as reliability and cycle time.  Generalization must be evaluated in a fair and convincing manner against state-of-the-art industry solutions, and not just straw-man ``classical'' approaches.
\end{enumerate}


These ``three E's of industrial reinforcement learning'' are critical for any method to be viable in real-world settings.
With these in mind, we present \textit{Super-Human InsErtion using Learning from Demonstration (SHIELD)}, a collection of design choices for the well-known DDPGfD algorithm which reflect several years of experimentation on industrial-insertion settings.
These choices are individually minor, but collectively produce a system that, to our knowledge, significantly outperforms any published learning approach to industrial insertion in terms of speed and robustness to variability in scene configuration.


The primary contributions of this paper can be summarized as follows:

\begin{enumerate}
    \item We conduct the first large scale systematic evaluation of an RL algorithm on an industrial robotic manipulation benchmark (NIST assembly board challenge), showing high reliability and robustness: 99.8\% out of 13K trials. 
    
    \item We describe in detail the SHIELD design choices which allow DDPGfD to attain these  results.

    \item We open-source a dataset containing all interaction data collected during our NIST experiments to support future work on machine learning for industrial manipulation.
    
    
    \item We present two challenge tasks and a comparison against \textit{human} performance (actual, not robot tele-operation) to demonstrate potential capabilities of this paradigm: a) HDMI cable insertion into a \textit{moving socket}, and b) blind key-insertion into a randomly-positioned lock, requiring searching ``in the dark''.  
    
\end{enumerate}
\label{introduction}
\section{PROBLEM STATEMENT}
\begin{figure}[thpb]
    \centering
   \includegraphics[width=9cm,keepaspectratio]{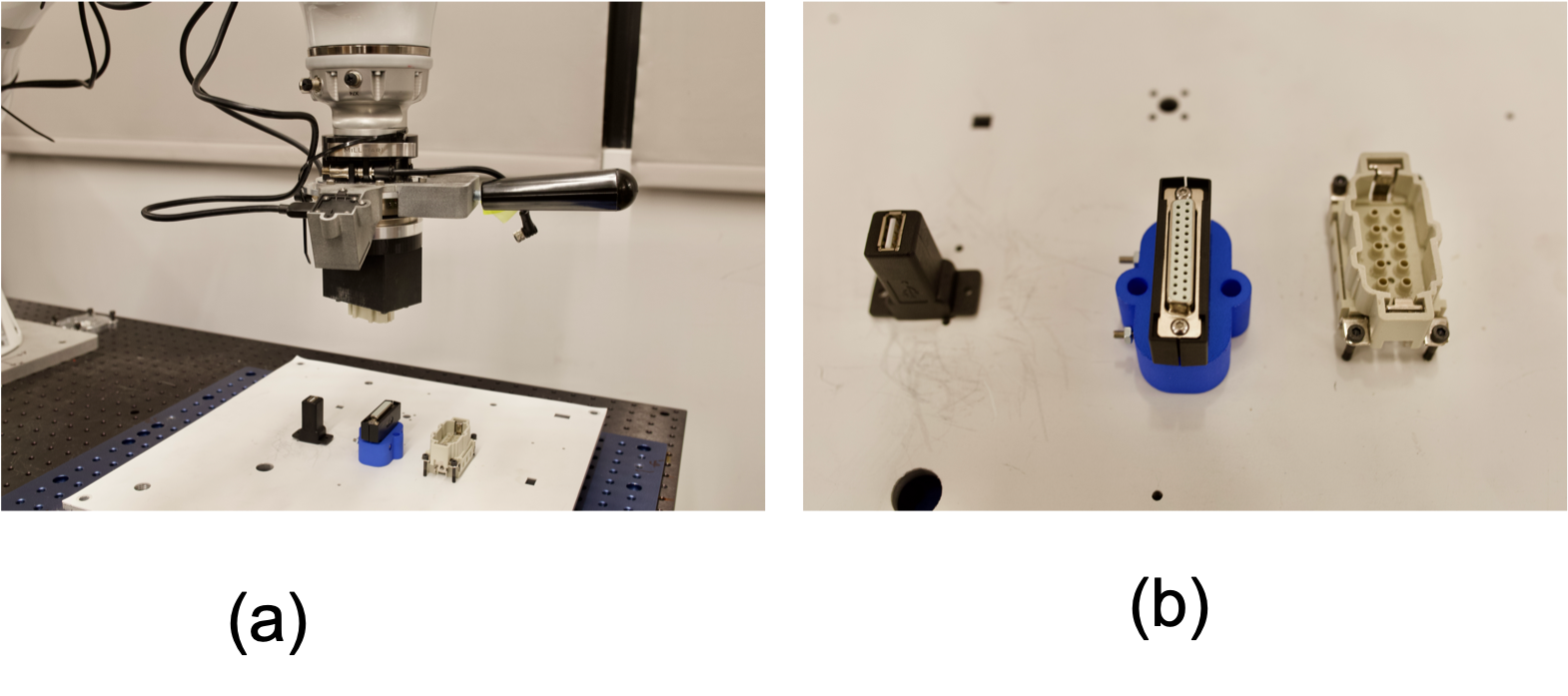}
    \caption{NIST robotic challenge board setup: (a) Robot inserting a connector, (b) Three representative connectors.}
    \label{nist}
\end{figure}

We aim to systematically evaluate our method in an industrial robotic benchmark and beyond.  To this end we consider three representative assembly tasks in the NIST assembly board as shown in Fig.\ref{nist}. In each of these tasks, the goal is to insert an industrial connector (\eg USB), into a corresponding socket.


To test the generalization capability of the learned polices, we ask the agent to perform connector insertions at new locations and orientations using the previously trained policies without any fine-tuning. For each of these tasks, the agent was trained with three different input modalities: 1)  proprioceptive information from the robot (Tool Center Point (TCP) pose and velocity) 2) proprioceptive information and wrist force/torque measurement 3) proprioceptive information, wrist force/torque measurement and wrist camera images. The agent is able to solve these tasks under all these modalities, and perform zero-shot transfer to new connector locations and orientations. In particular for policies trained with vision, we further test their generalization capabilities by randomly moving the NIST board by hand during the insertion process.

\begin{figure}
    \centering
    \includegraphics[width=9cm]{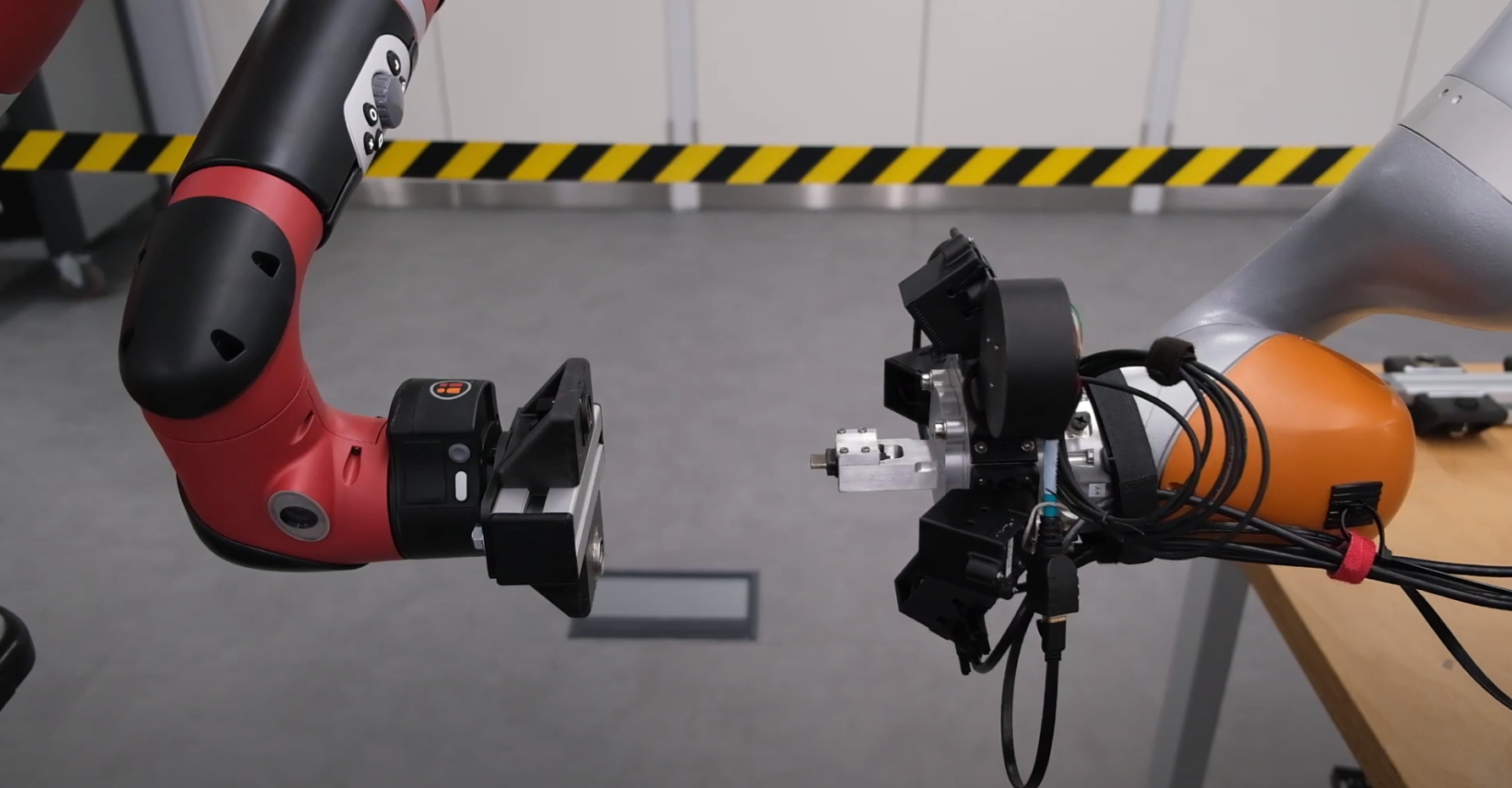}
    \caption{The cell setup for moving target HDMI insertion.}
    \label{fig:hdmi_setup}
\end{figure}

\begin{figure}
    \centering
    \includegraphics[width=9cm]{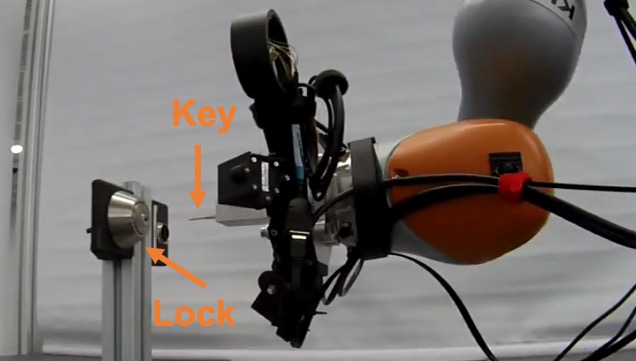}
    \caption{The cell setup for key insertion experiment.}
    \label{fig:loki_setup}
\end{figure}

 We also consider two more challenging tasks as shown in Fig.\ref{fig:hdmi_setup} and \ref{fig:loki_setup}. In the first task, the robot is asked to learn a policy to insert an HDMI connector into a socket that is held by another robot that moves along a randomized trajectory during the insertion process. In this task, the agent was not only expected to solve it reliably despite the moving target, but also to solve it as quickly as a human could, which is a critical threshold in the manufacturing industry. 

For the second task, we consider a key-lock insertion with situations where no visual information is available; these are typical scenarios in today's manufacturing sites.  The agent has to learn an appropriate searching strategy solely based on proprioceptive information and wrist force/torque measurement. This task is different and more difficult than the ones on the NIST board, which also use the same input modality: 1) the lock and key are more delicate pieces of hardware, and as they are made from metal they don't provide any material compliance; thus requiring even finer force-based motion to complete the task. 2) the starting position of the key is randomized and largely off the lock, which is not fully observable to the policy. In this case, the agent must learn to search for the keyhole position on the surface by maximum use of sensor readings.
\jsnote{we should emphasize that the whole purpose of this is to outperform spiral search in terms of speed and perturbation tolerance, to broaden the range of tasks that can be solved WITHOUT introducing cameras.}
\label{problem}
\section{RELATED WORK}




\cite{vecerik2019practical,vecerik2017leveraging} showed that it was possible to train a visual+force feedback controller from scratch to solve deformable connector insertion tasks in under 1 hr.  This method was built upon off-policy RL from demonstration (specifically DDPGfd) and used unsupervised pretrained visual features (VAE) to increase sample efficiency. The main contribution of this approach was to demonstrate that a purely learning-based solution to connector insertion was practical without relying on simulation.  However, results were only shown on synthetic 3D-printed parts, rather than real industrial connectors.


\cite{schoettler2020meta} introduces a sim2real approach that uses meta-learning to train an agent in simulation to quickly adapt to errors at (meta) test time, and demonstrated that this fast-adaptation ability transfers to the real-world. 
However, the setting was limited to 3-dim position-only actions, a 3mm workspace, and used a height-based reward function that only works in the immediate neighborhood of the socket.

\cite{luo2019reinforcement} describes a method which combines a model-based RL algorithm (iLQG) with an operational space force controller to solve high-precision insertion tasks. Using a guided policy search method they also train a neural network controller which is designed to explicitly consider force/torque measurements. 
The results show that this controller is able to generalize and solves peg-in-hole insertion tasks under unseen initial configurations of the prop positions.
However, this method does not demonstrate a very high reliability when generalizing to unseen peg-hole displacements.

\cite{johannink2019residual} introduces a method where a control policy is defined as the superposition of a user defined controller and a learned residual RL policy. The algorithm delegates the difficult parts of a task (such as contacts and external object dynamics) to a learned RL policy. This allows to approach problems that are common in manufacturing and are difficult to solve using conventional methods.
The algorithm is demonstrated on an insertion task of a block between two other freestanding blocks. The results show that the learned policy is able to complete insertions when the blocks are displaced and the scripted policy fails. However, they do not demonstrate high robustness and reliability and the selected task does not require the policy to be as precise as industrial connector insertion.

In principle any of these approaches is capable of competitive performance vs. scripted baselines. However, we pursue DDPGfD due to its comparative simplicity, which is an important factor when attempting to search the design space for our setting.

\label{related work}
\section{PRELIMINARIES}
We consider all tasks here that can be described as moving already-grasped  objects to their goal poses. We use a sparse reward system: 1) in the case without vision, the agent will get a binary reward at the termination of one trial episode: it will get a reward of one if its final position is within some tolerance of the target location, otherwise it will get a reward of zero; 2) in the case with vision inputs, we train an additional visual reward classifier to assign this binary reward. The goal of these tasks can be measured as maximizing such collected rewards. We make no particular assumptions about the encountered dynamics during tasks, especially during contacts. All dynamics need to be learned by the robot from interaction with the environment. Let $\bs_t$ and $\ba_t$ denote robot states and actions respectively; $\br(\bs_t,\ba_t)$ be the reward function related to the task, $T$ be the time horizon of a task.  Our problem can be formulated as finding
\begin{align*}
    &\max_{\ba_1,\ba_2 ...\ba_T}\sum_{t=1}^{T}\br(\bs_t, \ba_t) \\
    &s.t. \quad \bs_{t+1} = f(\bs_t,\ba_t) \,\, t=1,2...T-1,
\end{align*}
where $f$ governs the (unknown) system dynamics, and $\bs$, $\ba$ can also be subject to other algebraic constraints.\label{prelims}
\section{METHODS}
\label{sec:methods}
The workflow of our framework can be seen in Fig.\ref{cover_photo}, and can be described as follows:
\begin{itemize}
    \item Collect initial human demonstrations, saved to a replay buffer.
    \item (Optional) Train visual features on demonstration dataset.
    \item Roll out the current policy to collect more data.
    \item A human operator performs on-policy correction if necessary.
    \item (Optional) Add a learning curriculum.
    \item Actor and critic networks are trained asynchronously w.r.t. data collection.
\end{itemize}

\subsection{Reinforcement learning from demonstration}
RL methods are guided by reward signals. For contact-rich high-precision manipulation tasks as presented in this paper, the agent needs to perform sequenced fine-grained motion to achieve success. Using a shaped reward function to guide the RL agent requires a large amount of engineering effort to extract the necessary state information. On the other hand a pure sparse reward poses an extremely difficult exploration problem. 
Incorporating human demonstrations is a natural way to mitigate this issue \cite{nairdemo2018,vecerik2017leveraging,vecerik2019practical,dqfd,horgan2018distributed,Rajeswaran-RSS-18,fujimoto19a,dulacarnold2019challenges,zhu2018reinforcement,selfil}. A demonstrator can express intent by using a remote guiding device as in Fig.\ref{cover_photo}. This is an intuitive way of showing the agent possible solutions to the task, and can bootstrap a reasonable initial policy. We adopt and extend the Deep Deterministic Policy Gradient from Demonstration (DDPGfD) framework \cite{vecerik2017leveraging}, in that it has several convenient mechanism dealing with human demonstrations. The original DDPGfD algorithm and our modifications (in red) can be seen in Alg.\ref{alg:ddpgfd}. There are three modifications that significantly increase performance of the original DDPGfD algorithm: 

\begin{itemize}
    \item We remove the independent Gaussian noise added to the actor network -- thus we have a deterministic policy.
    \item The replay buffer retains all experience data and all transitions are weighted equally (no prioritization), actions from the human operator are regarded as the agent's own. 
    \item We also add task and action space curriculum to gradually increase the difficulty level of learning for the agent.
    \item We occasionally use the remote teaching device to override the agent's actions if it is not performing well, thus performing on-policy correction.
\end{itemize}

Our rationale for these changes is as follows. DDPG is a continuous variant of Deep Q learning, which is an off-policy RL algorithm, and it minimizes Bellman backup error as in Eq.\ref{bellman}:
\begin{equation}\label{bellman}
    L(\phi, \mathcal{D}) = \underset{(s,a,r,s^{\prime})\sim\mathcal{D}}{E}\bigg[\bigg(Q_{\phi}(s,a) - \Big(r + \gamma Q_{\phi}(s^{\prime}, \pi_{\theta}(s^{\prime}))\Big)\bigg)^2\bigg].
\end{equation}

Here, $\phi$ and $\theta$ are parameters for the Q network and policy network; replay buffer $\mathcal{D}$ contains data collected by an exploration policy; which is usually formed by adding exploration Gaussian noise (could be state dependent or not) to the acting policy.
We remove this Gaussian noise for two reasons: 1) exploration is addressed via demonstrations, and 2) i.i.d. random noise decreases the probability of successfully navigating narrow passages, which is precisely the problem we face in insertion tasks. This way we can easily reproduce behaviors from training in testing scenarios. In our experiments we found that the noise due to the stochastic learning process, coupled with the demonstration trajectories, provides sufficient exploration, despite the actor being deterministic. 

It is not uncommon to setup a prioritization mechanism in the replay buffer as seen in previous work \cite{HER}. 
However, in our regime we've found that prioritization schemes, e.g. using TD-error, consistently hurt performance.  We speculate that this is due to the highly off-policy regime in which we operate in RLfD -- not only does the replay buffer contain human demonstrations, but we also never evict samples, so early data becomes increasingly off-policy as training progresses.\footnote{See \cite{kumar2020discor} for a discussion on this effect.}
Instead, we always load all data seen so far into the replay buffer, and use uniform sampling to make sure that every piece of data gets utilized equally.
 
\begin{figure}[tbhp]
    \centering
    \includegraphics[width=8.5cm]{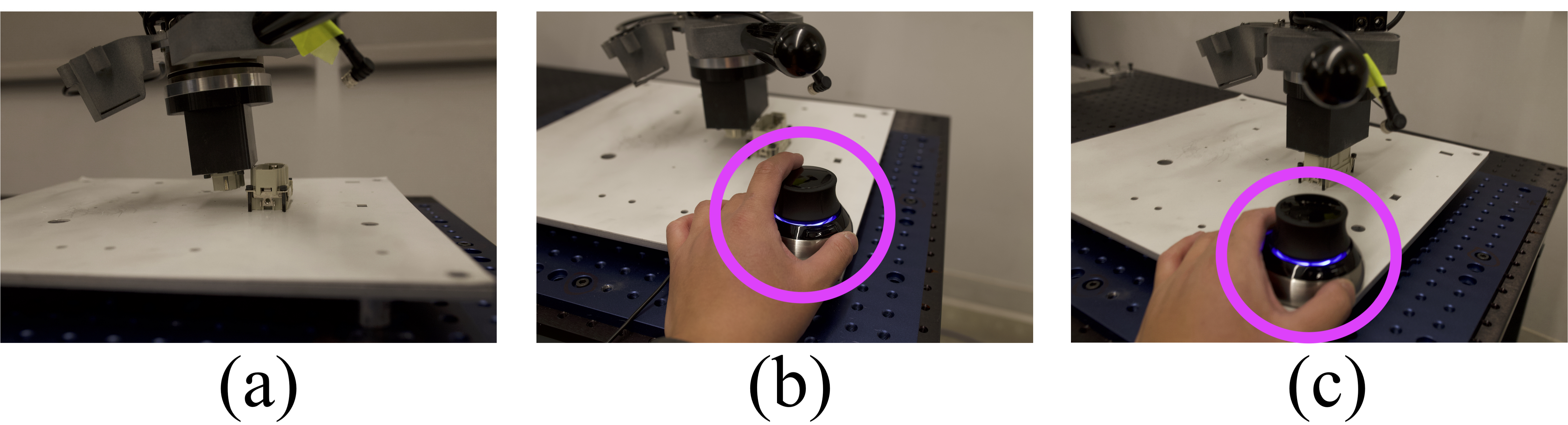}
    \caption{An illustration of on-policy correction: (a)the robot fails behind the sidewall of the male connector, (b)human engages by using remote teaching device, (c)human corrects robot's behavior}
    \label{correction}
\end{figure}

 The third modification we introduce is for a human operator to intervene and correct the RL agent if necessary. This ``on-policy'' correction has been used in previous works in imitation learning methods to correct distributional shift between the offline expert dataset and the online policy, and to incrementally train an agent to solve problems it encounters during learning\cite{dagger}. 
 This \textit{DAGGER} approach is particularly intuitive for insertion tasks.
 Consider the situation depicted in Fig.\ref{correction}(a) in which the plug has fallen to the outside of the socket.  
The correct strategy would be to lift up to avoid that local optimum and try moving to the top of the socket again to perform insertion. However, the agent has no mechanism of its own to develop this recovery behavior since the reward is sparse, and there is no local information guiding the agent to escape that local optimum. 
The human can simply intervene to guide the agent back into the region in which it has previously succeeded, and thanks to the off-policy structure of the algorithm the agent can incorporate this correction to future policies.
In practice, we found around ten such corrections would be enough for the agent to succeed in the task.
\begin{algorithm}[t]
     \caption{Robust DDPGfD}
     \label{alg:ddpgfd}
    \begin{algorithmic}[1]
  \Require  policy parameter $\theta$, target policy  parameter $\theta^{\prime}$,
            Q network parameter $\phi$, target Q network parameter $\phi^{\prime}$ 
\For {$t$ $\in$ [0, \emph{episode\_length}]}
    \State Collect demonstration, save to replay buffer $\mathcal{B}$
    \State Select an action $a_t = \pi_{\theta}(s_t)$ \st{$+ \sigma$, where $\sigma \sim \mathcal{N}(0, \Sigma)$}
    \State Take action $a_t$, observe$(s_t,a_t,s_{t+1},r_t)$, add it to $\mathcal{B}$
    \State \textcolor{red}{On-policy correction $a_t^{\prime}$ if necessary, overriding $a_t$}
    \State Compute TD target with minibatch data from $\mathcal{B}$ and target networks $Q_{\phi^{\prime}}$ and $\pi_{\theta^{\prime}}$
    \State Update $\theta$ and $pi$ with the TD target
    \State Update target networks $\theta^{\prime}$ and $\phi^{\prime}$ (Polyak averaging)
\EndFor
    \end{algorithmic}
\end{algorithm}

\subsection{Relative Coordinates and Goal Randomization}\label{relateive_rando}
In order to not over-fit to any particular absolute coordinate position, we introduce a mechanism allowing the agent to generalize to new locations. We drop all usage of pose information in the robot's base frame; instead we treat the robot's reset pose as the ``origin" for the task, and express all relevant proprioceptive information w.r.t. that origin.

From the agent's perspective this is equivalent to \textit{physically} moving the goal, but requires no additional mechanical apparatus beyond the robot itself.
A convenient consequence of this parameterization is that it is amenable to fine-grained control of the distribution of poses, which enabled the curriculum approach we describe in Section \ref{par:curriculum}.
We detail this procedure in the Alg.\ref{alg:relative}.  

\begin{algorithm}
     \caption{Relative coordinates and goal randomization}
     \label{alg:relative}
    \begin{algorithmic}[1]
     \Require perturbation magnitude $\epsilon$, nominal resetting pose $T_0$ w.r.t. robot base 
     \For {$i$ $\in$ [0, \emph{total\_episodes}]}
        \State Sample a noise $\sigma$ from uniform distribution $\mathcal{U}[-\epsilon, \epsilon]$
        \State Get randomized resetting pose $T_{i} = T_{0} + \sigma$ w.r.t. base
        \For{$t$ $\in$ [0, \emph{episode\_length}]}
        \State Get current pose $T_{i}^{t}$ w.r.t. base 
        \State Calculate relative pose $T_{rel}^{t} = T_{i}^{t} \times T_{i}^{-1}$
        \State Record  $T_{rel}^{t}$ to current state $s_{i}^{t}$
        \EndFor
     \EndFor
    \end{algorithmic}
\end{algorithm}

\subsection{Pre-trained Visual Features}
 As discussed previously, our goal is to train control policies in the real world in a timescale that is feasible for industrial use cases.  
 \cite{cabi2019scaling} demonstrated that it is possible to train a USB-insertion policy directly from pixels, but it took 8 hours to reach 80\% performance, which does not meet our requirements. Instead of training image features from reward, we advocate pre-training visual features using unsupervised learning objectives.  In this work used a variational auto-encoder (VAE), but there are numerous visual representation-learning approaches and this is an area of active research.  
 
 To gather this dataset we use the remote teaching device to jog the robot around the workspace to collect camera images to supplement the demonstration dataset. These images are necessary as the demonstration data typically only covers the part of the state-space seen by a near-optimal policy, which would lead to poor features when run outside this distribution.  This would be particularly problematic since we regress a human-labeled reward signal on the VAE features.
 
 The VAE training objective can be seen in Eq.\ref{eq:vae}:
\begin{equation} \label{eq:vae}
    \mathcal{L}(\theta,\phi;x) = \mathbb{E}_{q_{\phi}}(z|x)\Big[\text{log}p_{\theta}(x|z)\Big] - {\beta}D_{KL}\Big(q_{\phi}(z|x) || p(z)\Big).
\end{equation}

Here, $x$ are inputs to the VAE, which in our case are camera images; $z$ is the latent code, usually represented by a Gaussian distribution parameterized by mean and variance. The fist part in this loss function aims to reconstruct the original inputs; $p(z)$ in the second term of loss function is called the Prior that we often pick as a unit Gaussian distribution, the posterior $q_{\phi}(z|x)$ is asked to stay close with this prior distribution in the sense of some divergence measure.

After the VAE is trained, we fix the weights and discard the decoder. The encoder is used to extract latent codes from images, and these are passed to the agent. We'll detail the network architecture in the following section.

\subsection{Multi-modal policies}\label{multimodal}

\begin{figure}[thpb]
    \centering
   \includegraphics[width=9cm,keepaspectratio]{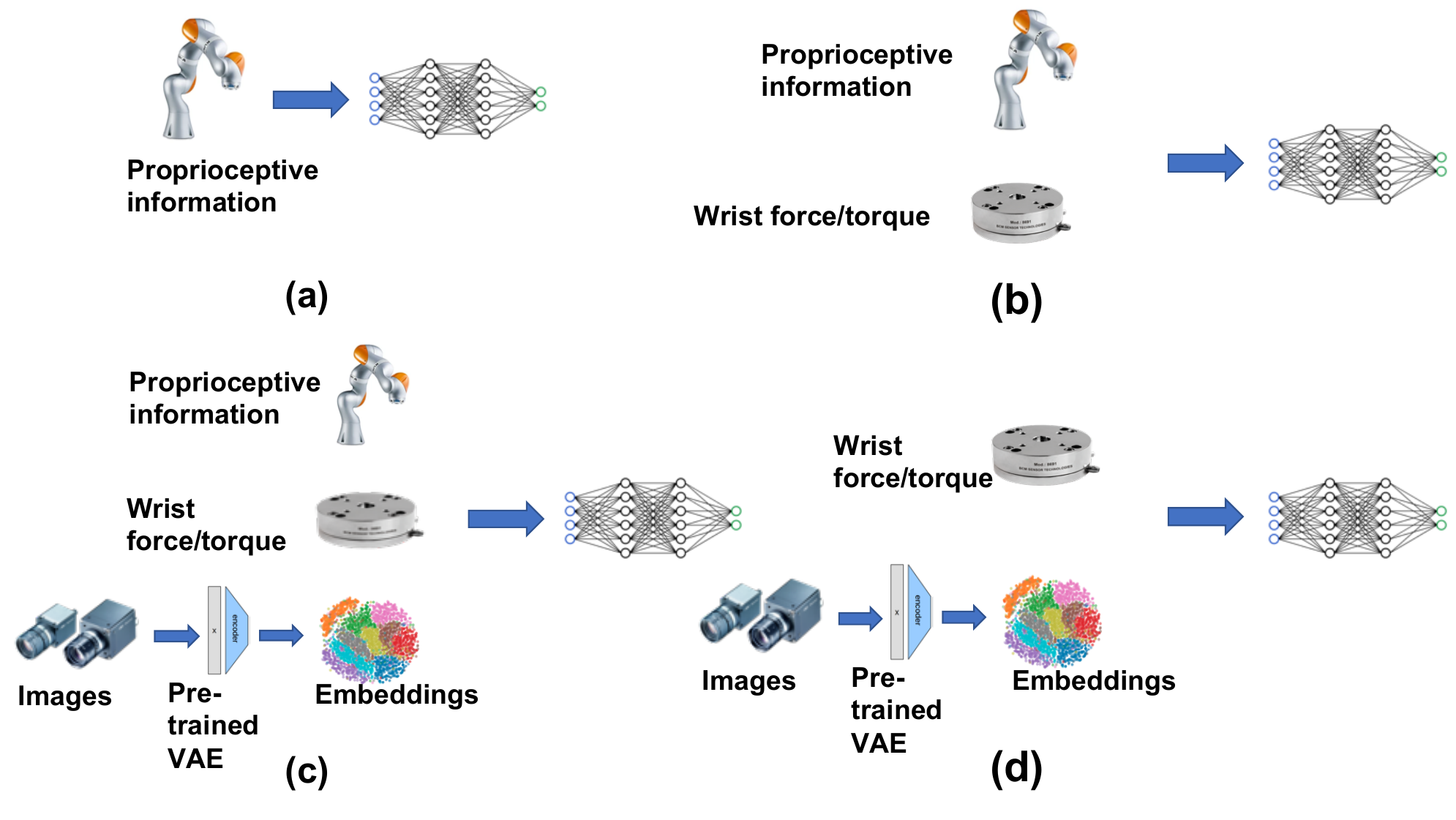}
    \caption{Different modality inputs to the policy.}
    \label{modality}
\end{figure}

We train and test the RL agent under four different sets of input modalities as in Fig.\ref{modality}. In Fig.\ref{modality}(a), the agent uses only the proprioceptive sensors of the robot arm. In Fig.\ref{modality}(b), we add wrist force/torque sensor on top of the proprioceptive information. In the Fig.\ref{modality}(c),(d), pre-trained visual features are made available to the agent; with the difference that, in (d), we rely purely on vision and force/torque, and the VAE is also used as reward detector 

\label{methods}
\section{EXPERIMENTS}
We report our experimental results in three sections: first the NIST board insertion tasks, which we evaluate with the first three input modalities as in Fig.\ref{modality}; second the moving target HDMI insertion, which we evaluate with the last input modality; last the key-lock insertion results, which doesn't use vision and performs long range ``in-the-dark search''.

\subsection{NIST board insertion results} \label{nist:subsection}
\subsubsection{Experiment setup}\label{nist:setup}
\begin{figure}[tbhp]
    \centering
    \includegraphics[width=8.9cm]{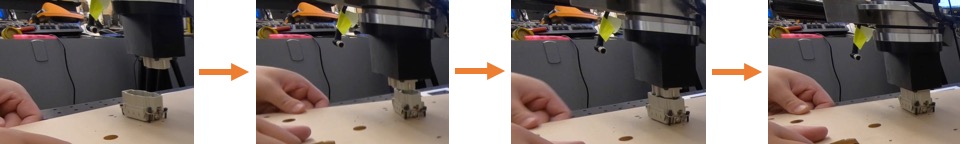}
    \caption{An illustration of dynamic the insertion task.}
    \label{dynamic}
\end{figure}

The experiment setup is shown in Fig.\ref{nist}. For each connector, we train policies with three different observation modalities: 1) proprioceptive information only, 2) proprioceptive information + wrist force/torque, 3) proprioceptive information + wrist force/torque + vision. For each trained policy, we evaluate its success rate on the trained location, with some perturbation noise added to the starting pose. We then move the female connector to a new location and 45 degree apart from its original orientation along the Z axis; and evaluate the success rate. To evaluate the generalization potential for vision-based polices, the agent performs insertion while a person is manually moving the NIST board as in Fig.\ref{dynamic}. This is a setting the agent was not trained on, but it verifies our hypothesis in Sec.\ref{relateive_rando}: the trained policy should generalize to any new condition as long as the relative transformation between start state and goal is within the range of training randomization. For all experiments, we use a Kuka iiwa arm, and control the end-effector Cartesian velocity using a customized impedance controller. 

We use a single Nvidia Titan RTX GPU for training. We use the ground truth pose to provide sparse reward label, i.e., if the current pose of the TCP is within some tolerance of target pose. Each episode lasts a maximum 10 seconds; early termination occurs on successful insertion or if the robot end-effector leaves the specified workspace. For vision-based experiments, we use a off-the-shelf USB camera, crop its image size to 320$\times$240.
\subsubsection{Baselines}
For a fair and convincing comparison, we contracted two independent robotic solution vendors to solve the same NIST board challenge. They used off-the-shelf tools and additional engineering on their side. We use their results as a baselines. We considered pure RL methods without demonstrations as an additional baseline, but none were able to solve these tasks and are not included in the discussion. Below we describe the vendors' approaches:

\paragraph{Vendor's approach} 
The robot moves to a predefined pose in the given female connector's frame where it's close to establishing contact. The robot then moves along the board's normal direction until an above-threshold force is detected. The robot regulates the force along the board normal direction at a predefined value and 0 torque along all 3 axes, meanwhile, searching in a spiral pattern in the board plane\cite{abb_manual}. The procedure stops when the movement along the board normal direction exceeds a threshold, i.e., the plug is considered aligned and is inserted into the hole.

\paragraph{Vendor's approach}
The nominal target female connector pose is defined such that the male connector and the female connector are perfectly aligned for insertion. Then we inject a 3D perturbation error: along the two axes in translation in parallel with the assembly board plane ($x, y$), and in rotation around the board's normal direction ($rz$). We start the insertion attempts using the nominal pose, repeating for 10 times with $rz$ rotation error smaller than 2 degrees. Then we increase the translation perturbation by 0.5mm along $x$ and/or $y$ axis following an outwards spiral pattern on a 2D grid, and repeat 10 attempts at each grid point. This process continues until consecutive failures are observed. 

\subsubsection{Results}
It is expected that vendors can solve these NIST assembly tasks with off-the-shelf tools. We are more interested in understanding how they solve these tasks and how well their solutions generalize. From task to task, the only change is the target pose; there is no additional engineering for each individual connector. On the vendors' side, a fair amount of time and effort is spent to engineer a solution and then fine-tune parameters for each connector type. More importantly, our methods are far more generalizable. They outperform vendor solutions by large margins in terms of perturbation ranges while keeping nearly 100\% success rates. This is a key performance indicator in semi-constrained environments with some level of uncertainties. Fig.\ref{table:dpgfd} shows the comparison of maximum perturbation of each connector. Table \ref{table:dpgfd} presents complete data for each connector under different testing conditions.

\begin{figure}[thpb]
    \centering
   \includegraphics[width=9cm,keepaspectratio]{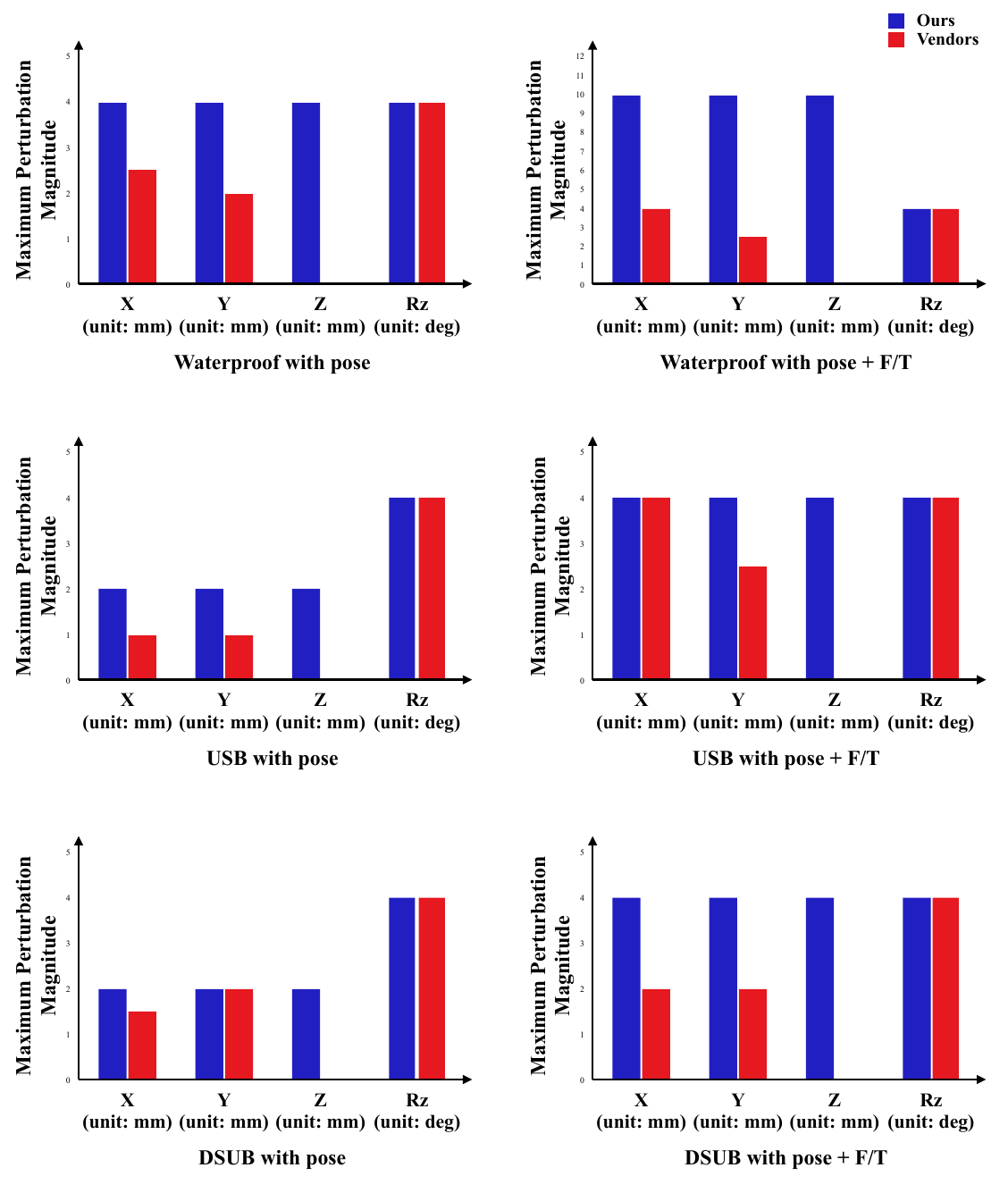}
    \caption{Comparison of maximum perturbation between our method and vendor's.} 
    \label{fig:vendor_comparison}
\end{figure}

\begin{table}
\begin{center}
\begin{adjustbox}{max width=9cm}
\begin{tabular}{ c | c | c | c}
 & Waterproof & USB & DSUB \\
 \hline
 pose + perturbation & 100\% (931/931) & 100\% (1357/1357) &  100\% (839/839)\\  [0.2cm]
 pose + 45 deg  & 100\% (1035/1035) & 99.4\% (623/627) &  100\% (960/960)\\  [0.2cm]
 pose + F/T + perturbation  & 100\% (801/801) & 98.9\% (791/800) &  100\% (1547/1547)\\  [0.2cm]
 pose + F/T + 45 deg  & 100\% (662/662) & 98.5\% (639/649) &  100\% (645/645)\\  [0.2cm]
 pose + F/T + vision + perturbation  & 100\% (653/653) & 100\% (802/802)&  100\% (788/788)\\[0.2cm]
 \multicolumn{4}{c}{\textbf{Total: 99.8\% (13069/13096)}}
\end{tabular}
\end{adjustbox}
\caption{Results of our method on three different connectors under different input modalities. ``perturbation" means that perturbations noise is added to its nominal resetting poses during training. ``45 deg" means that it was tested on a different location and rotated 45 degree apart around the Z axis.}
\label{table:dpgfd}
\end{center}
\end{table}

In total, we have performed 13096 trials, with a 99.8\% success rate. This strongly implies our method is robust and reliable. Most failures occurred with the USB connector, mainly due to the floating metal part inside the socket; active hard contact can cause changing dynamics. In the future we plan to look at ways to encourage a more gentle insertion policy.

\subsubsection{Dynamic insertion} As mentioned in Sec.\ref{nist:setup}, we test the vision-based policies in more challenging conditions. The agent needs to perform an insertion while a person is manually moving the NIST board as in Fig.\ref{dynamic}. This is a much harder task than static insertions, the agent's policy must be reactive so that it quickly adapts to changes in the environment. In essence this a closed-loop visual servoing policy, emerging from training using relative coordinates and goal randomization. Results show that our methods are able to solve these very challenging tasks with over 95\% success rate out of 50 trials. Most failures occur when the person generates high-frequency movement that is beyond the impedance controller's operating frequency, or far exceeding the training perturbation range. Complete videos can be seen on our website.

\subsection{Moving HDMI Insertion}
We have so far evaluated the robustness of our methods; another important metric in industrial robotics is cycle time. In this experiment, we evaluate the cycle time of our methods by asking the agent to do dynamic socket insertions with maximum possible speed. The goal is to demonstrate that the reliability and speed of our approach matches that of human hand-eye coordination.
\subsubsection{Experiment Setup}

The experiment involves the insertion of an HDMI plug into a dynamically moving socket. Socket motion is achieved by mounting it to a secondary robot arm. Neither the pose or velocity of the socket is given to the RL agent.
 
The HDMI socket is rigidly mounted to the flange of a 7-DOF Sawyer arm, the plug is rigidly attached to the flange of a 7-DOF Kuka IIWA robot. The IIWA has a wrist sensor providing Cartesian force-torque sensing, and a wrist mounted USB camera. At the start of each episode, the start position of the socket is randomized, the start pose of the plug is 15cm from the mean socket pose. During the episode the socket moves in a random circular motion in the plane perpendicular to the direction of insertion. In this case, we hide the pose information for the RL agent, it only has access to local TCP-frame velocity, wrist force/torque, and vision as in the fourth input modality in Fig.\ref{modality}. The RL agent is controlling the robot at 20 HZ.

\paragraph{Task Curriculum}
\label{par:curriculum}
To successfully train an RL agent for this challenging task, we employed a task curriculum. The task starts off with no randomization of the socket pose and no socket movement. As the agent improves, the task difficulty is increased, first by increasing the socket pose randomization, and then by increasing the movement speed of the socket. At the conclusion of the curriculum the socket initial position is randomized by up to +-6cm, and the movement speed is up to 2.5cm/s.
\jsnote{we should describe the schedule here or in an appendix}

\paragraph{Action Space Curriculum}
As we are aiming to minimize the insertion time, the agent needs to be able to request large velocities. However, an untrained agent may cause damage to the robot or environment. To improve training and safety we used a curriculum approach to the action space. Initially the available action space is quite small and conservative, but is grown exponentially (up to a maximum) as the agent policy becomes more reliable. 

\subsubsection{Agent Results}

We trained the RL agent on the task for 12 hours, with the replay buffer initialized with 25 successful demonstration trajectories. At the end of training we performed an evaluation run with the task randomization and agent action space set to maximum. We recorded the success/failure of each episode and the amount of time taken per episode. The agent had a 100\% insertion success rate over 1637 episodes, and a mean insertion time of 1093ms.

\begin{figure}[thpb]
    \centering
   \includegraphics[width=9cm,keepaspectratio]{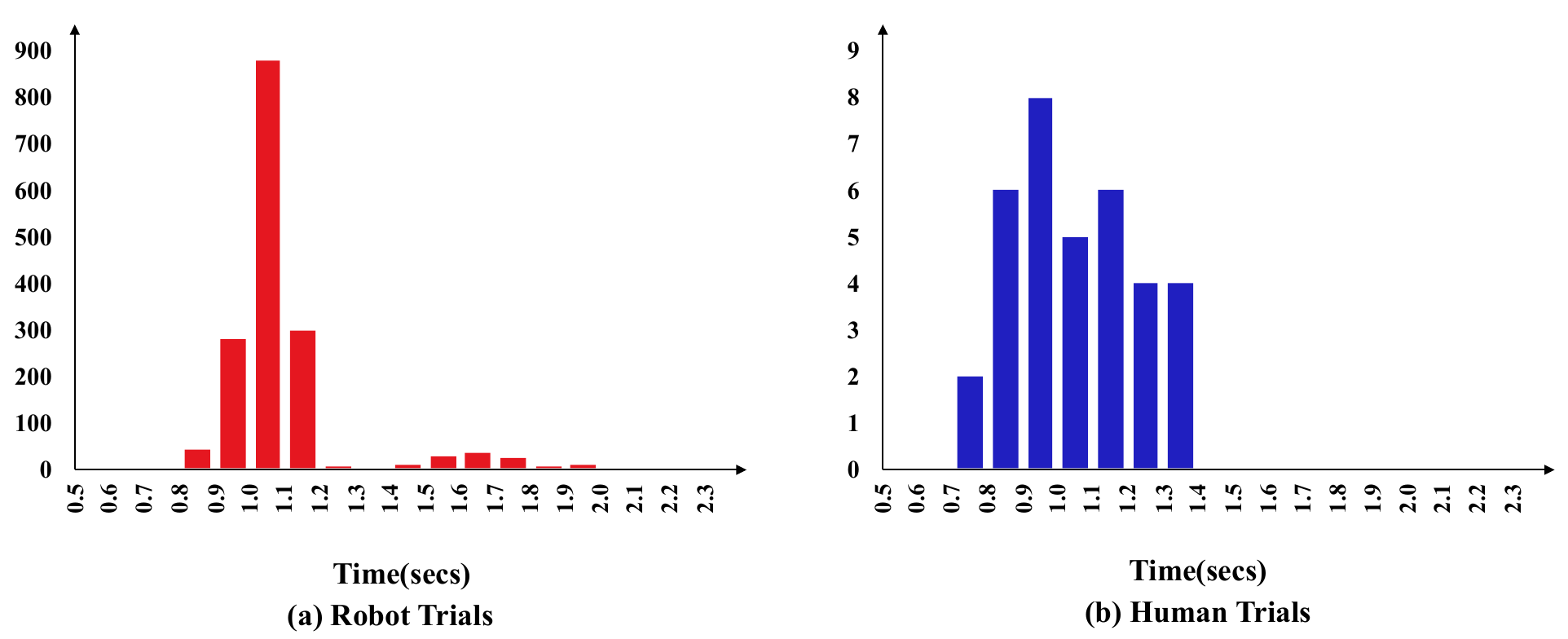}
    \caption{Insertion completion time comparison between our method and human}
    \label{fig:time_compare}
\end{figure}

\subsubsection{Human Results}

Our aim is to compare the RL agent performance to that of human hand-eye coordination. The setup of the human trials is similar to the RL agent, except the IIWA robot is removed and the human holds the HDMI plug in their hand. At the start of each episode the human participant is told to keep their their eyes closed and their hand positioned roughly 15cm from the mean position of the socket.
The robot holding the socket resets to a randomized position, and a beep is played over speakers to indicate the start of the episode. The human participant then opens their eyes and inserts the plug as fast as they can into the moving socket. We decided upon this setup to maximally match how the RL agent and the human participants experience the task. 
A total of 33 attempts were made by 2 human participants, with 100\% insertion success rate and a mean insertion time of 1041ms.

These experimental results show that the reliability and insertion time of the RL agent is comparable to that of human hand-eye coordination.

\subsection{Key-lock insertion}

As mentioned before, the plugs and sockets involved in previous insertion experiments were mainly industrial connectors, which have a particular range of sizes and material compliance characteristics. In this experiment we use a standard household metal lock and key as the plug and socket. 
The purpose of this experiment is to demonstrate that an RL agent can learn to insert objects with small surface area and clearance, as well as long range (when compared to the plug clearance) searching policies. The latter increases the range of perturbations than can be handled without a vision system when compared to industry standard search algorithms, which massively reduces system complexity.

\subsubsection{Experiment Setup}

This experiment consists of a key inserting into a lock with a RL-controlled robot. The key is attached to the end effector of 7-DOF Kuka IIWA robot while the lock is rigidly mounted in the cell. 

The robot is controlled by the RL agent at a rate of 20 Hz with a Cartesian velocity action space. The agent has access to full proprioceptive information and wrist force/torque as in the second modality in Fig.\ref{modality}; we also adopt the same relative coordinate and randomization mechanism as before. 

At the start of each episode the initial position of the robot is randomized. The key is always 5cm away from the lock but the position in x, y axis is sampled randomly (where z axis is aligned with the key). This offset is not observable since the TCP pose given to the agent is relative to the start of the episode. This makes the task more challenging: it forces the agent to learn a searching policy in order to find the lock opening and complete the insertion.

The episode terminates when the insertion is completed (which can be determined by the true TCP pose), the robot leaves the workspace, time limit of 15s is exceeded or the force/torque readings exceed a limit (to avoid damaging the key).

\paragraph{Starting pose randomization curriculum}

We employed a curriculum of the starting pose randomization. The offset magnitude in X/Y axis is sampled from a continuous uniform distribution between -$b$ and $b$. We start with $b = 0$ and slowly increase it as the agent is learning. This allowed the agent to learn insertion policies with up to 0.5cm offset in x and y axes, which the agent failed to do without the curriculum.

\subsubsection{Results}

The results are summarized in table \ref{table:loki_results_table}, measuring the success rate of the RL agent given different maximum starting pose offset. The ‘randomization’ column indicates maximum offset in X/Y axes that can be sampled at the start of the episode. These results show that our SHIELD agent is capable of learning insertion policies for objects of small clearance and surface area, as well as a long range search policy. The RL agent increases the range of a search policy to 5mm compared to vendors' results of 2mm in Section \ref{nist:subsection} Fig.\ref{fig:vendor_comparison}.

\begin{table}
\begin{center}
\begin{tabular}{ c | c }
 Maximum Pose Randomization & Insertion Success Rate \\ [0.5ex] 
 \hline
 0.1cm & 92\% (92 / 100) \\  
 0.3cm & 88\% (88 / 100) \\
 0.5cm & 87\% (87 / 100)
\end{tabular}
\caption{Key insertion results. The pose randomization value represents the maximum distance from the central position (right in front of the lock) to the sampled random positions in x,y axes (where z axis is aligned with the key)}
\label{table:loki_results_table}
\end{center}
\end{table}

\label{experiments}
\section{OPEN-SOURCE DATASET}
We colected a large amount of data during the NIST board evaluations, and we open sourced this dataset in the hopes of building interest in this setting within the machine-learning community.

The dataset consist of around 50 hours of robot interaction time across different tasks and input modalities; reward labels are also included. Further details can be found on our website:\url{https://sites.google.com/view/shield-nist}
\label{dataset}
\section{DISCUSSION AND FUTURE WORK}

We present a framework for learning robust robotic manipulation policies in industrial settings. We systematically evaluated the performance of our approach in the industrial robotic manipulation benchmark, NIST board. We show 99.8\% overall success rate out of 13K trials under a variety of different testing scenarios. The effectiveness of our approach is further demonstrated on two very challenging robotic tasks. The first is a moving target HDMI insertion task, the agent achieved a 100\% success rate out of 1637 trials and matched human performance in terms of cycle time. The second task is key-lock insertion, the agent learned a policy to successfully insert a metal key into a narrow clearance key-hole in the presence of initial condition randomization.

There are two key takeaways from our results: 
\begin{itemize}
    \item The NIST board assembly challenge represents a class of tasks solvable by robotic system integrators available today. However, these solutions require a semi-constrained environment and additional engineering effort for each individual assembly task, even if these are very similar to one another. Our approach mitigates these issues by leveraging demonstrations and training time randomization. Our approach can be seen as a ``general template" to solve a class of similar tasks with almost no parameter tuning except requiring a few demonstrations. Our learned policies outperform the traditional solutions by a large margin in terms of handling perturbations and generalization, verified by a large amount of real-world trials. This is essential to deploying any industrial robotic solution to the field.
    \item We also show that our method is able to unlock new applications for industrial robots. This can enable robots to enter unconstrained manufacturing environments and industries where the automation level remains low. 
\end{itemize}

We believe that our approach, together with the evaluation presented, would be the first step in applying learning-based robotic manipulation to the industrial robot field.

Future work can be regarded as following directions: 
\begin{itemize}
    \item Much larger-scale evaluation to verify the ``99.999\%" success rate, which can further convince the community of the tech-readiness of RL for industrial robotics.
    \item Better visual representations for cluttered scenes.  We found our vision model was not particularly robust to environmental changes such as lighting, background, distractions etc. VAE's loss function has a reconstruction term which attempts to model the entire scene, rather than focusing on the objects relevant to the task. As the objects become smaller in the visual field we find the performance of these agent decreases.
    \item We can further shorten the training time by leveraging recent advances in offline reinforcement learning plus online fine-tuning \cite{singh2020cog, nair2020accelerating, kumar2020conservative,levine2020offline}. 
\end{itemize}

\label{conclusions}

\printbibliography
\end{document}